\documentclass{article}

\usepackage{microtype}
\usepackage{graphicx}
\usepackage{subfigure}
\usepackage{amssymb}
\usepackage{booktabs} 
\usepackage{hyperref}

\usepackage[accepted]{icml2019}
\usepackage{amsmath}

\usepackage{color}

\newcommand{\cH}{\mathcal{H}}

\newcommand{\cN}{\mathcal{N}}

\definecolor{darkgreen}{rgb}{0,0.6,0}

\def\eq#1{\ref{#1}}

\newcommand{\be}{\begin{equation}}
\newcommand{\ee}{\end{equation}}
\newcommand{\bea}{\begin{equation} \begin{aligned}}
\newcommand{\eea}{\end{aligned} \end{equation}}

\newcommand{\bln}{\begin{align}}
\newcommand{\eln}{\end{align}}
\newcommand{\bst}{\begin{split}}
\newcommand{\est}{\end{split}}
\newcommand{\bi}{\begin{itemize}}
\newcommand{\ei}{\end{itemize}}
\newcommand{\ben}{\begin{enumerate}}
\newcommand{\een}{\end{enumerate}}

\newcommand{\bprop}{\begin{proposition}}
\newcommand{\eprop}{\end{proposition}}

\def\opt{{\textrm{opt}}}

\icmltitlerunning{The Effect of Network Width on Stochastic Gradient Descent and Generalization}

\begin{document}

\twocolumn[
\icmltitle{The Effect of Network Width on Stochastic Gradient Descent\\
and Generalization: an Empirical Study}

\begin{icmlauthorlist}
\icmlauthor{Daniel S. Park}{goog,aires}
\icmlauthor{Jascha Sohl-Dickstein}{goog}
\icmlauthor{Quoc V. Le}{goog}
\icmlauthor{Samuel L. Smith}{dm}
\end{icmlauthorlist}

\icmlaffiliation{goog}{Google Brain, Mountain View, USA}
\icmlaffiliation{dm}{DeepMind, London, UK}
\icmlaffiliation{aires}{Work done as a member of the Google AI Residency Program.}

\icmlcorrespondingauthor{Daniel S. Park}{danielspark@google.com}

\icmlkeywords{Machine Learning, SGD, stochastic, gradient, descent, batch size, learning rate, size, width, channels, langevin, optimal, ICML}

\vskip 0.3in
]

\printAffiliationsAndNotice{}

\begin{abstract}

We investigate how the final parameters found by stochastic gradient descent are influenced by over-parameterization. We generate families of models by increasing the number of channels in a base network, and then perform a large hyper-parameter search to study how the test error depends on learning rate, batch size, and network width. We find that the optimal SGD hyper-parameters are determined by a ``normalized noise scale,'' which is a function of the batch size, learning rate, and initialization conditions. In the absence of batch normalization, the optimal normalized noise scale is directly proportional to width. Wider networks, with their higher optimal noise scale, also achieve higher test accuracy. These observations hold for MLPs, ConvNets, and ResNets, and for two different parameterization schemes (``Standard'' and ``NTK''). We observe a similar trend with batch normalization for ResNets. Surprisingly, since the largest stable learning rate is bounded, the largest batch size consistent with the optimal normalized noise scale decreases as the width increases.

\end{abstract}

\section{Introduction}
\label{s:introduction}

Generalization is a fundamental concept in machine learning, but it remains poorly understood \cite{ZhangBHRV16}.
Theoretical generalization bounds are usually too loose for practical tasks \cite{harvey2017nearly,neyshabur2017pac,bartlett2017spectrally, dziugaite2017computing, zhou2018non,nagarajan2018deterministic}, and practical approaches to hyper-parameter optimization are often developed in an ad-hoc fashion \cite{sculley2018iclr}. A number of authors have observed that Stochastic Gradient Descent (SGD) can be a surprisingly effective regularizer \citep{keskar2016large, wilson2017marginal, sagun2017empirical, SGDBinference, smithLe,SGDVinference, soudry2018implicit}. In this paper, we provide a rigorous empirical study of the relationship between generalization and SGD, which focuses on how both the optimal SGD hyper-parameters and the final test accuracy depend on the network width.

This is a broad topic, so we restrict the scope of our investigation to ensure we can collect thorough and unambiguous experimental results within a reasonable (though still substantial) compute budget. We consider training a variety of neural networks on classification tasks using SGD without learning rate decay (``constant SGD''), both with and without batch normalization \citep{batchnorm}. We define the performance of a network by its average test accuracy at ``late times" and over multiple training runs. The set of optimal hyperparameters $\cH_\opt$ denote the hyper-parameters for which this average test accuracy was maximized. We stress that optimality is defined purely in terms of the performance of the trained network on the test set. This should be distinguished from references to ideal learning rates in the literature, which are often defined as
the learning rate which converges fastest during training \cite{efficient,fishermeanfield}. Our use of optimality should also not be confused with references to optimality or criticality in some recent studies \citep{minibatch, largebatch} where these terms are defined with respect to efficiency of training, rather than final performance.

Given these definitions, we study the optimal
hyper-parameters and final test accuracy of networks in the same class but with different widths. Two networks are in the same class if one can be obtained from the other by adjusting the numbers of channels. For example, all
three-layer perceptrons are in the same class, while a two-layer perceptron is in a different class to a three-layer perceptron. For simplicity, we consider a ``linear family" of networks,
\be
\{ \cN_{w_1},~ \cdots~,~\cN_{w_k} \},
\ee
each of which is obtained from a base network $\cN_1$ by introducing a widening
factor, much in the spirit of wide residual networks \cite{WideRN}.
That is, network $\cN_{w}$ can be obtained from $\cN_1$ by widening every layer by a constant factor of $w$. We aim to identify a predictive relationship between the optimal hyper-parameters $\cH_\opt$ and the widening factor $w$. We also seek to understand the relationship between network width and final test accuracy.

We will find that a crucial factor governing both relationships is the ``normalized noise scale''. As observed in \cite{SGDBinference,SGDVinference,threefactors, smithLe}, and reviewed in section \ref{ss:nsgd},
when the learning rate is sufficiently small,
the behaviour of SGD is determined by the noise scale $g$, where for SGD,
\be\label{gdef 1}
g = {\epsilon N_\textrm{train} \over B},
\ee
and for SGD with momentum,
\be\label{gdef 2}
g = {\epsilon N_\textrm{train} \over B(1-m)}.
\ee
Here, $\epsilon$ is the learning rate,
$B$ is the batch size, $m$ is the momentum coefficient, and
$N_\textrm{train}$ is the size of the training set. \citet{smithLe} showed that there is an optimal noise scale $g_{\textrm{opt}}$, and that any setting of the hyper-parameters satisfying $g = g_{\textrm{opt}}$ will achieve optimal performance at late times, so long as the effective learning rate $\epsilon/(1-m)$ is sufficiently small. We provide additional empirical evidence for this claim in section \ref{s:results}. However in this work we argue that to properly define the noise introduced by SGD, $g$ should be divided by the square of a weight scale. A quick way to motivate
this is through dimensional analysis. In a single SGD step, the parameter update is proportional to the learning rate multiplied by the gradient. Assigning the parameters units of $[\textrm{weight}]$, and the loss units of $[\textrm{loss}]$, the gradient has units of $[\textrm{loss}]/[\textrm{weight}]$. This implies that the learning rate has dimensions of $[\textrm{weight}]^2/ [\textrm{loss}]$. The scale of the loss is controlled by the choice of cost function and the dataset. However the weight scale can vary substantially over different models in the same class. We hypothesize that this weight scale is controlled by the scale of the weights at initialization, and it will therefore depend on the choice of parameterization scheme \cite{NTK}.
Since the noise scale is proportional to the learning rate, we should therefore divide it by the square of this weight scale. 

In this work we will consider two parameterization schemes, both defined in section \ref{ss:NTK}. In the ``standard'' scheme most commonly used in deep learning, the weights are initialized from an isotropic Gaussian distribution whose standard deviation is inversely proportional to the square root of the network width. As detailed above, this work will consider families of networks obtained by multiplying the width of every hidden layer in some base network $\cN_1$ by a multiplicative factor $w$. Thus the normalized noise scale,
\be\label{std barg}
\bar{g}(\cN_w) = {g \over (\sigma_0 /\sqrt{w})^2}
= {g w \over \sigma_0^2} \quad
\textrm{for standard scheme}.
\ee
The standard deviation $\sigma_0$ defines the weight scale of the base network, which for our purposes is just a constant. An alternative parameterization was recently proposed, commonly referred to as ``Neural Tangent Kernel'' parameterization, or ``NTK'' \cite{NTKL2, NTKGAN1, NTK, NTKGAN2}. In this scheme, the weights are initialized from a Gaussian distribution whose standard deviation is constant, while the pre-activations are multiplied by the initialization factor \cite{Glorot, He} {\it after} applying the weights to the activations in the previous layer. Since the weight scale in this scheme is independent of the widening factor,
\be\label{NTK barg}
\bar{g} = {g / \sigma_0^2} \quad
\textrm{for NTK scheme}.
\ee
By finding the optimal normalized noise scale for families of
wide residual networks (WRNs), convolutional networks (CNNs)
and multi-layer perceptrons (MLPs) for image classification tasks on
CIFAR-10 \cite{cifar10}, Fashion-MNIST (F-MNIST) \cite{fmnist} and MNIST 
\cite{mnist}, we are able to make the following
observations:
\begin{enumerate}
\item Without batch normalization, the optimal normalized noise is proportional to the widening factor.
That is,
\be\label{main result}
\bar{g}_{\textrm{opt}} (\cN_w) \propto w.
\ee
See section \ref{s:results} for plots. This result implies,
\bi
\item For the standard scheme, the optimal
value of $\epsilon/B$ stays constant with
respect to the widening factor.
\item For the NTK scheme, the optimal value
of $\epsilon/B$ is proportional to the widening factor.
\ei
\item The definition of the noise scale does not apply to
networks with batch normalization, since the gradients of individual examples depend on the rest of the batch. However we have observed that the trend expressed in
equation
\eq{main result} still holds in a weaker sense. Considering networks parameterized using the NTK scheme,
\bi
\item When the batch size is fixed, the optimal learning rate increases
with the widening factor.
\item When the learning rate is fixed, the optimal batch size decreases
with the widening factor.
\ei
\end{enumerate}

Residual networks \citep{he2016deep} obey the trend implied by equation \ref{main result} both with and without batch normalization. Furthermore for all networks, both with and without batch normalization, wider networks consistently perform better on the test set \citep{neyshabur2018towards, lee2018deep}. 

The largest stable learning rate is proportional to $1/w$ in the standard scheme, while it is constant for the NTK scheme (discussed further in section \ref{ss:NTK}). This implies that the largest batch size consistent with equation \ref{main result} decreases as the network width rises. Since the batch size cannot be smaller than one, these bounds imply that there is a critical network width above which equation \eq{main result} cannot be satisfied. 

The paper is structured as follows. In section \ref{s:background}, we
review the background material and introduce our notation.
In section \ref{s:experiments} we describe how the experiments were performed, while the empirical results are presented in section \ref{s:results}. In section \ref{s:comments} we discuss our findings and their implications.

\section{Background}
\label{s:background}

\subsection{Standard vs. NTK Parameterization Schemes}
\label{ss:NTK}
In the standard scheme,
the pre-activations $z^{l+1}_i$ of layer $(l+1)$ are related to
the activations $y^l_j$ of layer $l$ by
\be
z^{l+1}_i = \sum_{j=1}^{n_l} W^l_{ij} y^l_j 
+ b^{l+1}_i,
\ee
and weights and biases are initialized according to
\be\label{He init}
W^l_{ij} \sim \cN\left( 0,~ {\sigma_0^2 \over n_l} \right), \quad
b^{l+1}_i = 0.
\ee
The scalar $n_l$ denotes the input dimension of
the weight matrix, and $\sigma_0^2$ is 
a common weight scale shared across all models in the same class.
For fully connected layers $n_l$ is the dimension of the input, while for convolutional layers $n_l$
is the filter size multiplied by the number of input channels.
By inspecting equation \eq{He init}, we can see that the weight scale is inversely proportional to the square root of the widening factor $w$. Following the discussion in the introduction, we arrive at equation \eq{std barg} by normalizing the learning rate accordingly:
\be
\bar{\epsilon}(\cN_w) = {\epsilon \over (\sigma_0 /\sqrt{w})^2}
= {\epsilon w \over \sigma_0^2} \quad
\textrm{for standard scheme}.
\ee
Meanwhile in the NTK scheme \cite{NTKL2, NTK},
the pre-activations $z^{l+1}_i$ are related to the activations $y^l_j$ of the previous layer by,
\be
z^{l+1}_i = {1 \over \sqrt{n_l}} \left( \sum_{j=1}^{n_l} W^l_{ij} y^l_j \right) + \beta_l b^{l+1}_i ,
\ee
and weights and biases are initialized according to
\be
W^l_{ij} \sim \cN(0, \sigma_0^2), \quad
b^{l+1}_i = 0.
\ee
Notice that the scaling factor ${1 \over \sqrt{n_l}}$ is introduced after applying the weights, while the parameter $\beta_l$ controls the effect of bias. We set $\beta_l = 1/\sqrt{n_l}$ in all experiments. The weight scale is independent of the widening factor $w$, leading to a normalized learning rate which also does not depend on $w$,
\be
\bar{\epsilon} = {\epsilon / \sigma_0^2} \quad
\textrm{for NTK scheme}.
\ee
We therefore arrive at the normalized noise scale of equation \eq{NTK barg}. The test set performance of NTK and standard networks are compared in section \ref{ap:ntk vs std} of the supplemental material.

The learning rate has an upper-bound defined by convergence criteria and numerical stability. This upper-bound will also be on the order of the square of the weight scale, which implies that the upper-bound for $\bar{\epsilon}$ is approximately
constant with respect to the widening factor.
It follows that the stability bound for the bare learning rate $\epsilon$
scales like $1/w$ for the standard scheme \cite{fishermeanfield}, while it remains constant for the NTK scheme. We provide empirical evidence supporting these stability bounds in section \ref{ap:stability} of the supplementary material. A major advantage of the NTK parameterization is that we can fix a single learning rate and use it to train an entire family of networks $\{\cN_w \}$ without encountering 
numerical instabilities. We therefore run the bulk of our experiments using the NTK scheme.

\subsection{Noise in SGD}
\label{ss:nsgd}

\citet{smithLe} showed that for SGD and SGD with momentum, if the effective learning rate is sufficiently small the dynamics of SGD are controlled solely by the noise scale (equations \eq{gdef 1} and \eq{gdef 2}). This implies that
the set of hyperparameters for which the network achieves maximal
performance at ``late times" is well approximated by a level
set of $g$ (i.e., the set of hyperparameters for which
$g = g_\textrm{opt}$). We define ``late times'' to mean sufficiently long for the validation accuracy to equilibrate. To verify this claim, in our experiments we will make two independent measurements of $g_\textrm{opt}$. One is obtained by holding the learning rate fixed and sweeping over the batch size, while the other is obtained by holding the batch size fixed and sweeping over the learning rate. We find that these two measures of $g_\textrm{opt}$ agree closely, and they obtain the same optimal test performance. We refer the reader to section \ref{ss:training time} for further discussion of training time, and section \ref{ap:grid} of the supplementary material for experiments comparing the test set performance of an MLP across a two dimensional grid of learning rates and batch sizes.

This analysis breaks down when the learning rate is too large \cite{yaida2018fluctuation}. However empirically for typical batch sizes (e.g., $B \lesssim 1000$ on ImageNet), the optimal learning rate which maximizes the test set accuracy is within the range where the noise scale holds\footnote{i.e., linear scaling of $\epsilon$ and $B$ does not degrade performance.} \cite{goyal2017accurate, smith2017dont, largebatch, minibatch}. Our experiments will demonstrate that this does not contradict the common observation that, at fixed batch size, the test set accuracy drops rapidly above the optimal learning rate.

When a network is trained with batch normalization, the gradients for individual samples depend on the rest of the batch, breaking the analysis of \citet{smithLe}. Batch normalization also changes the gradient scale. We therefore do not expect equation \eq{main result} to hold when batch normalization is introduced. However we note that at fixed batch size, the SGD noise scale is still proportional to the learning rate.

\section{Experiments}
\label{s:experiments}

\subsection{Overview}
\label{ss:overview}

We run experiments by taking a linear family of networks, and finding
the optimal normalized noise scale for each network on a given task.
We measure the optimal noise scale in two independent ways---we
either fix the learning rate and vary the batch size, or fix the
batch size and vary the learning rate. We use fixed Nesterov momentum $m=0.9$.

We first describe our experiments at fixed learning rate.
We train 20 randomly initialized networks for each model in the family
at a range of batch sizes, and compute the test accuracy after training ``sufficiently long'' (section \ref{ss:training time}). We then compute
the average trained network performance and find the batch size $B$ with the best average
performance $\mu_B$. The ``trained performance'' refers to the average test accuracy of ``trained runs'' (runs whose final test accuracy exceeds $0.2$). We compute the standard deviation $\sigma_B$ of
the trained accuracy at this batch size and find all contiguous batch
sizes to $B$ whose average accuracy is above $\mu_B - 2 \sigma_B /\sqrt{n_B}$, where $n_B$ is the number of trained runs at batch size $B$. This procedure selects all batch sizes whose average accuracy is within two standard error deviations of $\mu_B$, and it defines the ``best batch size interval'' $[B_0, B_1]$, from which we compute the ``best normalized noise scale interval'' $[\bar{g}_1, \bar{g}_0]$. We estimate the optimal normalized noise scale by $\bar{g}_\textrm{opt} = (\bar{g}_0 + \bar{g}_1)/2$. When $B_0 \neq B_1$, we include an error bar
to indicate the range. The procedure for computing the optimal normalized noise scale in
experiments with fixed batch size $B$ is analogous to the procedure above; we train all networks 20 times for a range of learning rates and compute the best learning rate interval $[\epsilon_0, \epsilon_1]$.

Our main result is obtained by plotting the optimal normalized noise scale $\bar{g}_\textrm{opt}(\cN_w)$ against the widening factor $w$ (in the absence of batch normalization). When batch normalization is introduced, the definition of the noise scale is not valid. In this case, we simply report the optimal inverse batch size (learning rate) observed when fixing the learning rate (batch size), respectively.
To make the plots comparable,
we rerun the estimation procedure used for finding the optimal value of $\bar{g}$
when estimating the optimal value of $1/B$ (batch size search) and
$\epsilon$ (learning rate search).

\subsection{Training Time}
\label{ss:training time}

To probe the asymptotic ``late time'' behaviour of SGD, we run our experiments with a very large compute budget,
where we enforce a lower bound on the training time both
in terms of the number of training epochs and the number of parameter updates.
When we run learning rate searches with fixed batch size,
we take a reference learning rate, for which the training steps
are computed based on the epoch/step constraints, and then scale this
reference training time accordingly for different learning rates. 
Although we find consistent relationships between the batch size, learning rate, and test error, it is still possible that our experiments are not probing asymptotic behavior \citep{minibatch}.

We terminate early any training run whose test accuracy falls below $0.2$ at any time $t$ beyond $20\%$
of the total training time $T$. We verified that at least 15 training runs completed successfully for each experiment (learning rate/batch size pair). See sections \ref{ap:ttime def} and \ref{ap:configs} of the supplementary material for a detailed description of the procedure used to set training steps and impose lower bounds on training time.

\subsection{Networks and Datasets}
\label{ss:parameters}

We consider three classes of networks; multi-layer perceptrons,
convolutional neural networks and residual networks. We use
ReLU nonlinearities in all networks, with softmax readout.
The weights are initialized at criticality with $\sigma_0^2 =2$
\citep{He, deepprop}.

We perform experiments on MLPs, CNNs and ResNets. We consider MLPs with 1, 2 or 3 hidden layers and denote the $d$-layered perceptron with uniform width $w$ by the label $d\textrm{LP}_w$. Our family of convolutional networks $\textrm{CNN}_w$ is
obtained from the celebrated LeNet-5 (figure 2 of \citet{lenet})
by scaling all the channels, as well as the fully connected layers, by
a widening factor of $w/2$. Our family of residual networks
$\textrm{WRN}_w$ is obtained from table 1 of \citet{WideRN} by taking $N=2$ and $k=w$. Batch normalization is only explored
for CNNs and WRNs. 

We train these networks for classification tasks on
MNIST, Fashion-MNIST (F-MNIST), and CIFAR-10.
More details about the networks and datasets used in the experiments can be
found in section \ref{ap:networks and datasets} of the supplementary material.

We train with a constant learning rate, and do not consider data augmentation, weight decay or other regularizers in the main text. Learning rate schedules and regularizers introduce additional hyper-parameters which would need to be tuned for every network width, minibatch size, and learning rate. This would have been impractical given our computational resources. However we selected a subset of common regularizers (label smoothing, data augmentation and dropout) and ran batch size search experiments for training WRNs on CIFAR-10 in the standard parameterization with commonly used hyper-parameter values. We found that equation \ref{main result} still held in these experiments, and increasing network width improved the test accuracy. These results can be found in section \ref{ap:reg} of the supplementary material.

\begin{figure*}[ht!]
  \centering
  \begin{tabular}{cc|cc}
  \includegraphics[height=3.00cm]{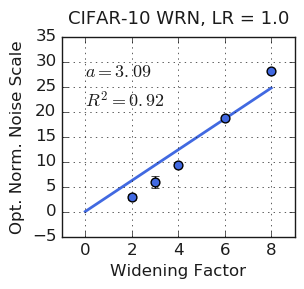} &
  \includegraphics[height=3.00cm]{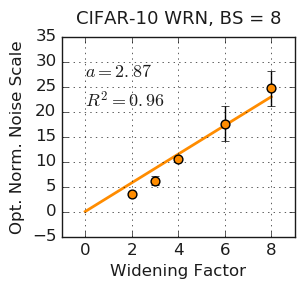} &
  \includegraphics[height=3.00cm]{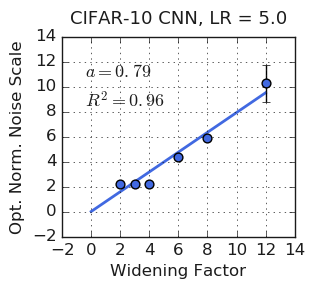} &
  \includegraphics[height=3.00cm]{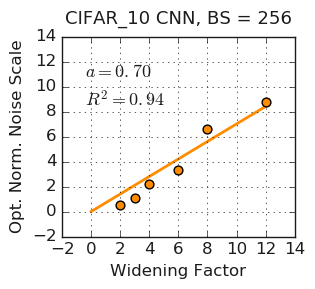} \\
  \includegraphics[height=3.00cm]{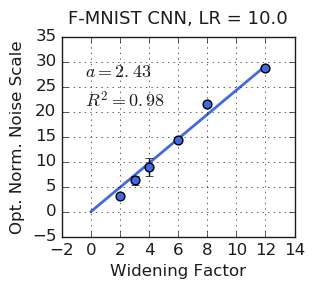} &
  \includegraphics[height=3.00cm]{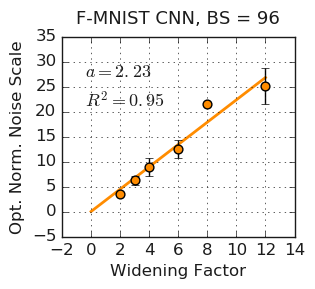} &
  \includegraphics[height=3.00cm]{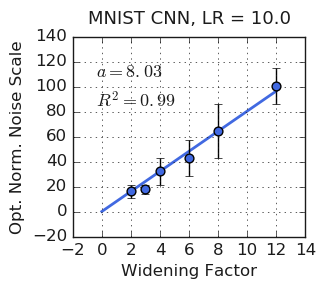} &
  \includegraphics[height=3.00cm]{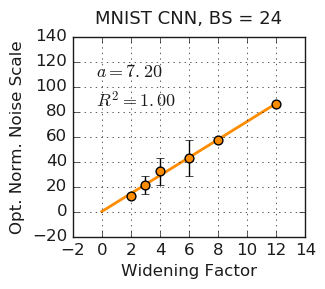} \\
  \includegraphics[height=3.00cm]{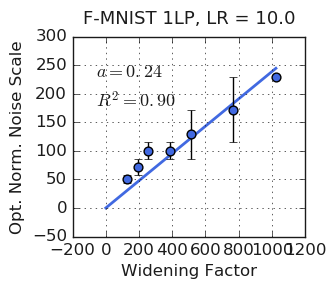} &
  \includegraphics[height=3.00cm]{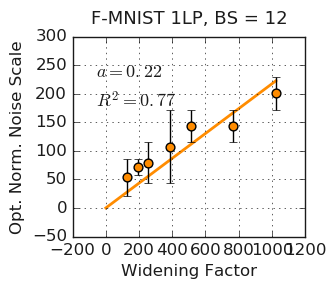} &
  \includegraphics[height=3.00cm]{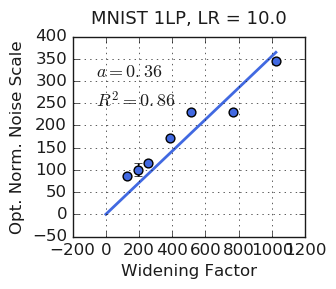} &
  \includegraphics[height=3.00cm]{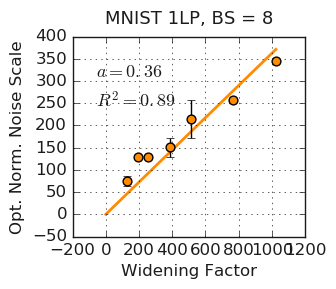} \\
  \includegraphics[height=3.00cm]{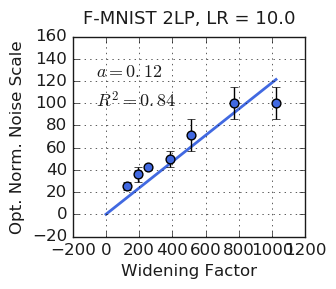} &
  \includegraphics[height=3.00cm]{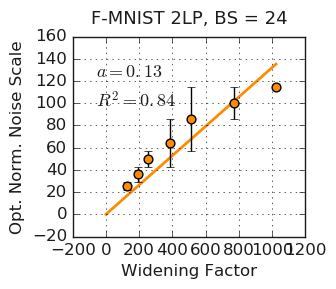} &
  \includegraphics[height=3.00cm]{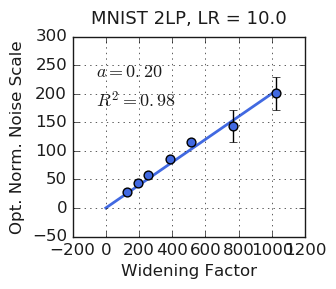} &
  \includegraphics[height=3.00cm]{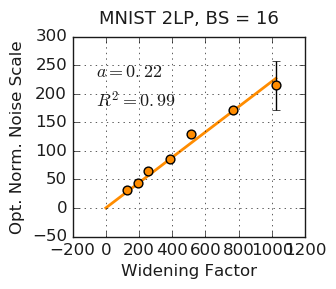} \\
  \includegraphics[height=3.00cm]{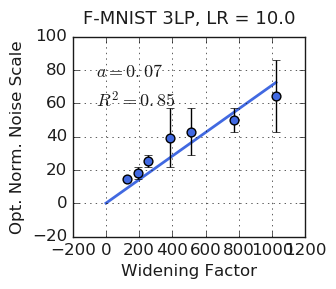} &
  \includegraphics[height=3.00cm]{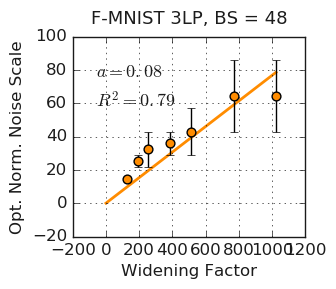} &
  \includegraphics[height=3.00cm]{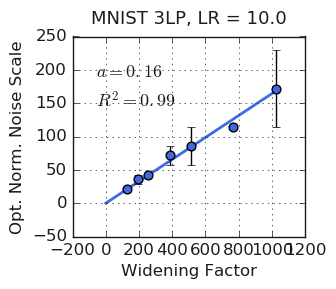} &
  \includegraphics[height=3.00cm]{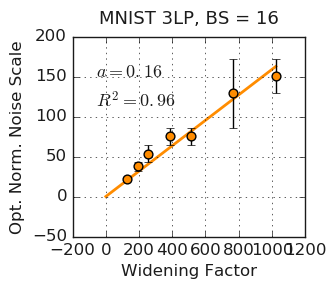}
  \end{tabular}
  \vskip - 0.1in
  \caption{Optimal normalized noise scale vs. widening factor for NTK parameterized networks trained without batch normalization. The optimal normalized noise scale
  (y-axis) has units of 1000/[loss] while the widening
  factor (x-axis) is unitless. The proportionality constant $a$ for each plot has the same units as the optimal normalized noise scale.
  This optimal normalized noise scale is obtained via batch size search
  for the blue plots, and via learning rate search for the orange plots.
  The fixed learning rate or batch size used to generate each plot is indicated in the title. Every dataset-network-experiment tuple exhibits a clear linear relationship as predicted by equation \ref{main result}.
}
  \label{f:nobx}
  \vskip -0.1in
\end{figure*}

\section{Experiment Results}
\label{s:results}

Most of our experiments are run with networks parameterized in the
NTK scheme without batch normalization. These experiments, described in section \ref{ss:nobx}, provide the strongest evidence for our main result, $\bar{g}_\textrm{opt} (\cN_w) \propto w$ (equation \eq{main result}). We independently optimize both the batch size at constant learning rate and the learning rate at constant batch size, and confirm that both procedures predict the same optimal normalized noise scale and achieve the same test accuracy.

We have conducted experiments on select dataset-network pairs
with standard parameterization and without batch normalization in section \ref{ss:nxbx}. In this section we only perform batch size search at fixed learning rate. Finally we run experiments on NTK parameterized WRN and CNN networks
with batch normalization in section \ref{ss:nobo}, for which we perform both batch size search and learning rate search.
Some additional batch size search experiments for WRNs parameterized in the standard scheme with batch normalization can be found in section \ref{ap:stdbn} of the supplementary material, while we provide an empirical comparison of the test performance of standard and NTK parameterized networks in section \ref{ap:ntk vs std}. We provide a limited set of experiments with additional regularization in section \ref{ap:reg} of the supplementary material.

In section \ref{ss:new}, we study how the final test accuracy depends on the network width and the normalized noise scale.

\subsection{NTK without Batch Normalization}
\label{ss:nobx}

In figure \ref{f:nobx}, we plot the optimal normalized noise scale
against the widening factor for a wide range of network families.
The blue plots were obtained by batch size search with fixed learning rate, while the orange
plots were obtained by learning rate search with fixed batch size.
The fixed batch size or learning rate is given in the title of each
plot alongside the dataset and network family. As explained in section \ref{ss:overview}, the error
bars indicate the range of normalized noise scales
that yield average test accuracy within the $95$\% confidence interval
of the best average test accuracy. For each plot we fit the proportionality constant $a$ to the equation $
\bar{g}_\textrm{opt} (\cN_w) = aw$,
and provide both $a$ and the $R^2$ value. We observe a good fit in each plot. The proportionality constant $a$ is computed  independently for each dataset-network family
pair by both batch size search and learning rate search, and these two constants consistently agree well.

\begin{figure}[t!]
  \centering
  \includegraphics[height=4.3cm]{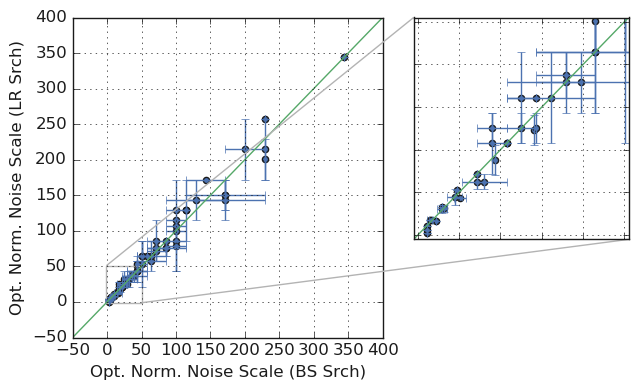}
  \vskip -0.1in
  \caption{The optimal normalized noise scale obtained via
  learning rate search plotted against that obtained via batch size search
  for all experiments with NTK parameterized networks without batch normalization.
  The optimal normalized noise scale is given in units of $10^3$/[loss].
  The green line is the line $y=x$.}
  \label{f:bs vs lr}
  \vskip -0.15in
\end{figure}

We can verify the validity of our assumption
that the set of hyper-parameters yielding optimal performance is given by a level set of $\bar{g}$,
by comparing both the optimal normalized noise scale $\bar{g}_\textrm{opt}$ and the maximum test
set accuracy obtained by batch size
search and learning rate search.
The obtained values for both search methods for all experiments
have been plotted against each other
in figures \ref{f:bs vs lr} and \ref{f:hth_perf}.
Further evidence for our assumption can be found in section \ref{ap:grid}
of the supplementary material.
Finally, for each triplet of dataset-network-experiment type, we have plotted
the test set accuracy against the scanned parameter (either batch size
or learning rate) in figure \ref{f:nobx search}
of section \ref{ap:plots}.

\begin{figure}[t]
  \centering
  \begin{tabular}{cc}
  \includegraphics[height=3.1cm]{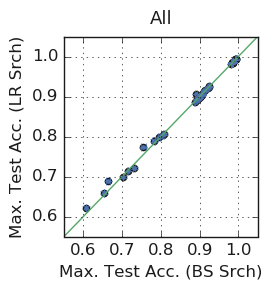} &
  \includegraphics[height=3.1cm]{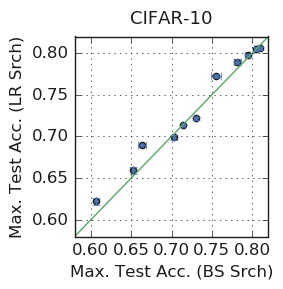} \\
  \includegraphics[height=3.1cm]{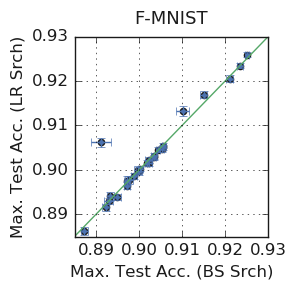} &
  \includegraphics[height=3.1cm]{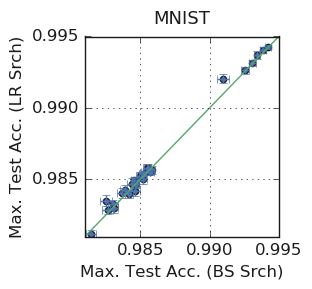}
  \end{tabular}
  \vskip -0.1in
  \caption{The maximum performance of networks obtained via
  learning rate search plotted against that obtained via batch size search
  for all experiments with NTK parameterized networks without batch normalization.
  The green line is the line $y=x$.}
  \label{f:hth_perf}
  \vskip -0.15in
\end{figure}

\begin{figure*}[t!]
  \centering
  \begin{tabular}{ccccc}
  \includegraphics[height=2.6cm]{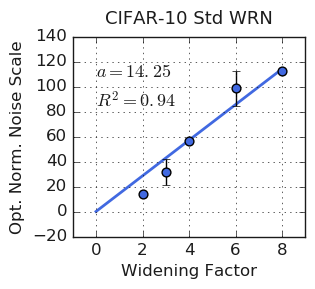} &
  \includegraphics[height=2.6cm]{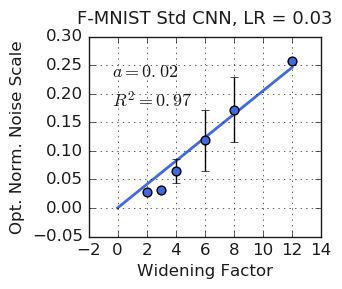} &
  \includegraphics[height=2.6cm]{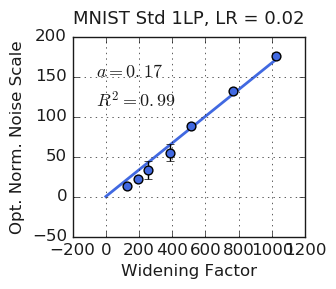} &
  \includegraphics[height=2.6cm]{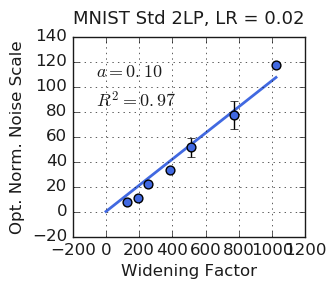} &
  \includegraphics[height=2.6cm]{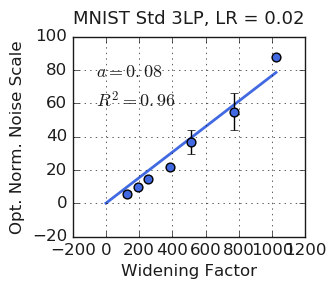}
  \end{tabular}
  \vskip -0.1in
  \caption{The optimal normalized noise scale vs. widening factor for networks parameterized using the standard scheme without batch normalization.
  The optimal normalized noise scale and the proportionality constant $a$ are given in units of $10^6$/[loss], and
  the optimal normalized noise scale is obtained via batch size search. All five plots exhibit a clear linear relationship.}
  \label{f:nxbx}
  \vskip -0.1in
\end{figure*}

\begin{figure}[t!]
  \vskip -0.05in
  \centering
  \begin{tabular}{ll}
  \includegraphics[height=3.0cm]{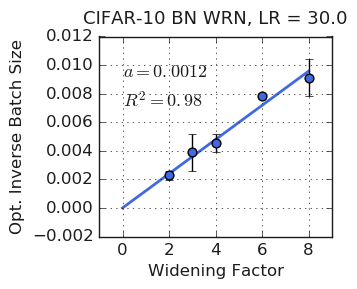} &
  \includegraphics[height=3.0cm]{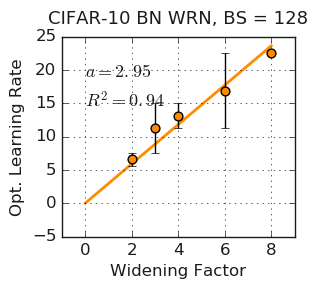} \\
  \includegraphics[height=3.0cm]{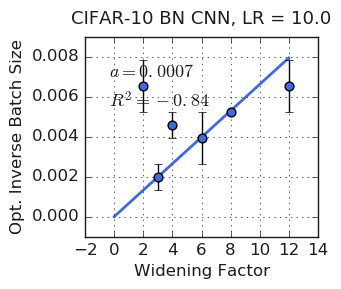} &
  \includegraphics[height=3.0cm]{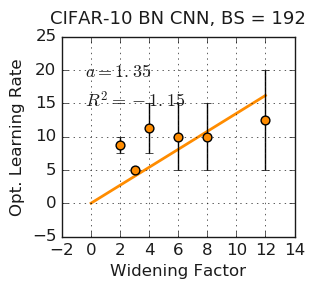} \\
  \includegraphics[height=3.0cm]{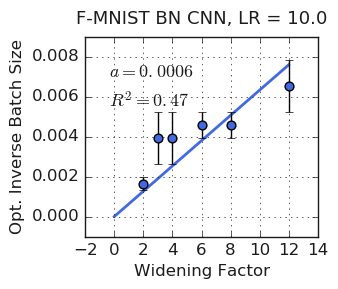} &
  \includegraphics[height=3.0cm]{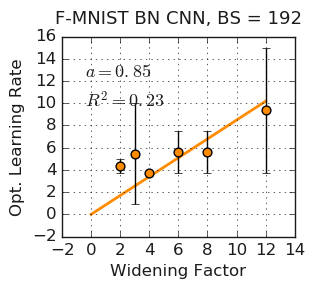} \\
  \includegraphics[height=3.0cm]{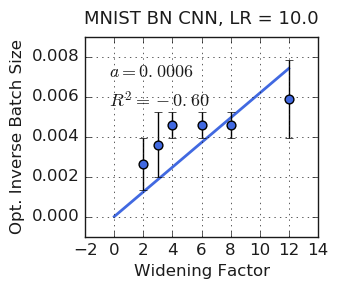} &
  \includegraphics[height=3.0cm]{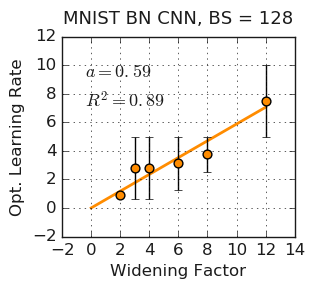} 
  \end{tabular}
  \vskip -0.1in
  \caption{The optimal inverse batch size/learning rate vs.
  widening factor for NTK parameterized networks
  with batch normalization.}
  \label{f:nobo}
  \vskip -0.15in
\end{figure}

\begin{figure}[t]
  \vskip -0.05in
  \centering
  \begin{tabular}{cc}
  \includegraphics[height=3.0cm]{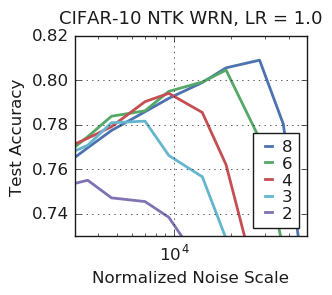} &
  \includegraphics[height=3.0cm]{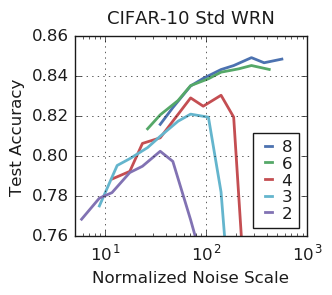} \\
  \includegraphics[height=3.0cm]{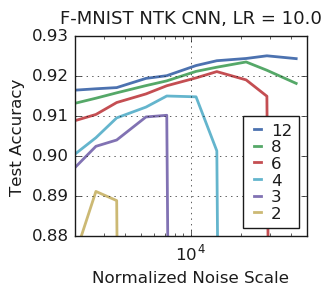} &
  \includegraphics[height=3.0cm]{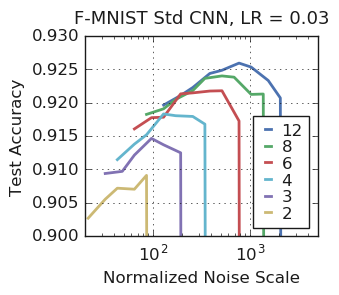} \\
  \includegraphics[height=3.0cm]{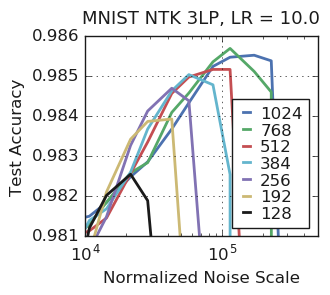} &
  \includegraphics[height=3.0cm]{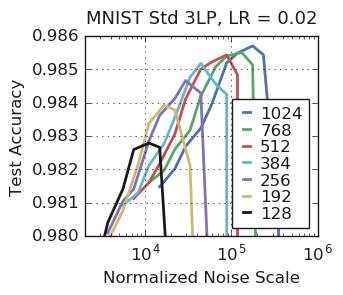} \\
  \end{tabular}
  \vskip -0.1in
  \caption{
  The test set accuracy plotted against normalized noise scale (in units of [loss]$^{-1}$)
  from experiments with fixed learning rate without batch normalization.
  The datasets, networks, parameterization schemes, and learning
  rates are indicated in the plot titles, except for the second plot where
  separate learning rates were used for different networks (see section \ref{ss:nxbx}).
  The legend indicates the widening factor, and the x-axis is in log-scale.
  }
  \label{f:acc vs noise}
  \vskip -0.15in
\end{figure}

\subsection{Standard without Batch Normalization}
\label{ss:nxbx}

Due to resource constraints, we have only conducted experiments for select
dataset-network pairs when using the standard parameterization
scheme, shown in figure \ref{f:nxbx}. The optimal normalized noise scale is found using batch size search only. For CIFAR-10 on WRN, networks with $w=2,3,4$
were trained with the learning rate $0.0025$, while networks
with $w=6,8$ were trained with a smaller learning rate of $0.00125$
due to numerical instabilities (causing high failure rates for wide models when $\epsilon = 0.0025$). Once again, we observe a clear linear relationship between the optimal normalized noise scale and the widening factor.
We provide additional experiments incorporating data augmentation, dropout and label smoothing in section \ref{ap:reg} of the supplementary material, which also show the same linear relationship.

\subsection{NTK with Batch Normalization}
\label{ss:nobo}

We have only conducted experiments with batch normalization for families of wide residual networks and CNNs. We perform both batch size search and learning rate search.
The results of these experiments
are summarized in figure \ref{f:nobo}.
Unlike the previous sets of experiments, we do not report
an optimal normalized noise scale,
as this term is poorly defined when batch normalization is used. Rather, we report the optimal
inverse batch size (learning rate) for the given fixed learning
rate (batch size) respectively. However as discussed previously,
when the batch size is fixed, 
the SGD noise 
is still proportional to the learning rate. 
A clear linear trend is still present for wide residual networks, however this trend is much weaker in the case of convolutional networks.

\subsection{Generalization and Network Width}
\label{ss:new}

We showed in section \ref{ss:nobx} that in the absence of batch normalization, the test accuracy of a network of a given width is determined by its normalized noise scale. Therefore in figure \ref{f:acc vs noise}, we plot the test accuracy as a function of noise scale for a range of widths. We include a variety of networks trained without batch normalization with both standard and NTK parameterizations. In all cases the best observed test accuracy increases with width. This is consistent with previous work showing that test accuracy improves with increasing over-parameterization \cite{neyshabur2018towards,lee2018deep,novak2018sensitivity}. See section \ref{ap:plots} of the supplementary material for plots of test accuracy in terms of the raw learning rate and batch size instead of noise scale.

More surprisingly, the dominant factor in the improvement of test accuracy with width is usually the increased optimal normalized noise scale of wider networks. To see this, we note that for a given width the test accuracy often improves slowly below the optimum noise scale and then falls rapidly above it. Wider networks have larger optimal noise scales, and this enables their test accuracies to rise higher before they drop. Crucially, when trained at a fixed noise scale below the optimum, the test accuracy is very similar across networks of different widths, and wider networks do not consistently outperform narrower ones. This suggests the empirical performance of wide over-parameterized networks is closely associated with the implicit regularization of SGD.

\begin{figure}[t!]
  \centering
  \begin{tabular}{cc}
  \includegraphics[height=3.0cm]{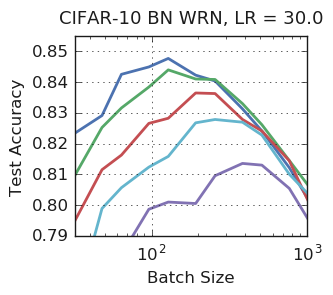} &
  \includegraphics[height=3.0cm]{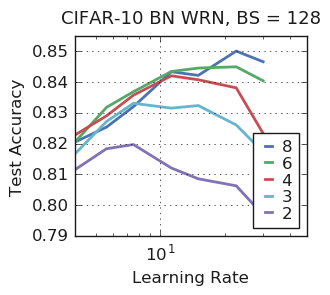} \\
  \includegraphics[height=3.0cm]{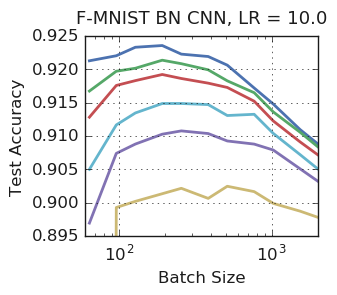} &
  \includegraphics[height=3.0cm]{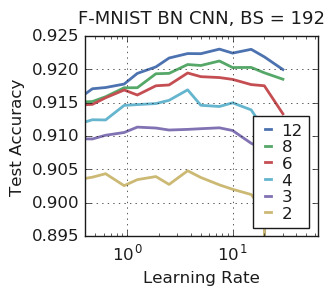}
  \end{tabular}
  \vskip -0.1in
  \caption{The test accuracy plotted against batch size/learning rate
  for experiments with fixed learning rate/batch size respectively.
  The networks are NTK-parameterized and use batch normalization.
  The x-axis is log-scaled, and the legend indicates the widening factor of the plotted networks.
  The dataset, network architecture and values for the fixed parameters are indicated in the title.}
  \label{f:nobo scans}
  \vskip -0.15in 
\end{figure}

In figure \ref{f:nobo scans} we examine the relationship between generalization and network width for experiments with batch normalization. Since the normalized noise scale is not well-defined, we plot the test accuracy for a range of widths as a function of both the batch size at fixed learning rate, and the learning rate at fixed batch size. We provide plots for WRNs on CIFAR-10 and CNNs on F-MNIST, both in the NTK parameterization. Once again, we find that the best observed test accuracy consistently increases with width. However the qualitative structure of the data differs substantially for the two architectures. In the case of WRNs, we observe similar trends both with and without batch normalization. In the NTK parameterization, wider networks have larger optimal learning rates (smaller optimal batch sizes), and this is the dominant factor behind their improved performance on the test set. For comparison, see figure \ref{f:nobx search} in section \ref{ap:plots} of the supplementary material for equivalent plots without batch normalization. However in the case of CNNs the behaviour is markedly different, wider networks perform better across a wide range of learning rates and batch sizes. This is consistent with our earlier observation that WRNs obey equation \ref{main result} with batch normalization, while CNNs do not.

\section{Discussion}
\label{s:comments}

\textbf{Speculations on main result:}
The proportionality relation (equation \eq{main result}) between the optimal
normalized noise scale and network width holds remarkably robustly
in all of our experiments without batch normalization. This relationship provides a simple prescription which predicts how to tune SGD hyper-parameters as width increases. We do not have a theoretical explanation for this phenomenon. However intuitively it appears that noise enhances the final test performance, while the amount of noise a network can tolerate is proportional to the network width. Wider networks tolerate more noise, and thus achieve higher test accuracies. Why SGD noise enhances final performance remains a mystery \citep{keskar2016large, sagun2017empirical, SGDBinference, SGDVinference,smithLe}.

\textbf{Implications for very wide networks:}
As noted in the introduction, the largest batch size consistent with equation \ref{main result} decreases with width (when training with SGD + momentum without regularization or batch normalization). To clarify this point, we consider NTK networks and standard networks
separately. For NTK networks, the learning rate can stay constant
with respect to the width without introducing numerical instabilities.
As the network gets wider equation \eq{main result} requires $B \propto \epsilon / w$, which
forces the batch size of wide networks to have a small value to achieve
optimality. For standard networks, $B \propto \epsilon$.
However in this case the learning rate $\epsilon \propto 1/w$ must decay as the width increases in order for the SGD to
remain stable \citep{fishermeanfield}. We provide empirical evidence for these stability bounds in section \ref{ap:stability} of the supplementary material. In both cases the batch size must eventually be reduced if we wish to maintain optimal performance as width increases.

Unfortunately we have not yet been able to perform additional experiments at larger widths. 
However if the trends above hold for arbitrary widths, then there would be a surprising implication. Since the batch size is bounded from below by one, and the normalized learning rate is bounded above by some value $\bar{\epsilon}_\textrm{max}$ due to numerical stability, there is a maximum noise scale we can achieve experimentally. Meanwhile the optimal noise scale increases proportional to the network width. This suggests there may be a critical width for each network family, at which the optimal noise scale exceeds the maximum noise scale, and beyond which the test accuracy does not improve as the width increases, unless additional regularization methods are introduced.

\textbf{Comments on batch normalization:}
The analysis of small learning rate SGD proposed by \citet{smithLe} does not hold with batch normalization, and we therefore anticipated that networks trained using batch normalization might show a different trend. Surprisingly, we found in practice that residual networks trained using batch normalization do follow the trend implied by equation \ref{main result}, while convolutional networks trained with batch normalization do not.

\textbf{Conclusion:}
We introduce the normalized noise scale, which extends the analysis of small learning rate SGD proposed by \citet{smithLe} to account for the choice of parameterization scheme. We provide convincing empirical evidence that, in the absence of batch normalization, the normalized noise scale which maximizes the test set accuracy is proportional to the network width. We also find that wider networks perform better on the test set. A similar trend holds with batch normalization for residual networks, but not for convolutional networks. We consider two parameterization schemes and three model families including MLPs, ConvNets and ResNets. Since the largest stable learning rate is bounded, the largest batch size consistent with the optimal noise scale decreases as the width increases.

\section*{Acknowledgements}
We thank Yasaman Bahri, Soham De, Boris Hanin, Simon Kornblith, Jaehoon Lee, Luke Metz, Roman Novak,
George Philipp, Ben Poole, Chris Shallue, Ola Spyra, Olga Wichrowska and Sho Yaida for helpful discussions.

\bibliography{ref}
\bibliographystyle{icml2019}

\clearpage

\onecolumn
\icmltitle{Supplementary Material}

\appendix

\section{Networks and Datasets}
\label{ap:networks and datasets}

We consider MLPs with 1, 2 or 3 hidden layers. Each layer has the same number of hidden units, $w$. We denote the $d$-layered perceptron with width $w$ by the label $d\textrm{LP}_w$. We do not consider batch normalization for these networks, and we consider the range of widths,
\be
w \in \{ 128, 192, 256, 384, 512, 768, 1024 \}.
\ee

We consider a family of convolutional networks $\textrm{CNN}_w$,
obtained from LeNet-5 (figure 2 of \citet{lenet})
by scaling all the channels, as well as the widths of the fully connected layers, by
a widening factor of $w/2$ (the factor of ${1 \over 2}$ allows integer $w$ for all experiments). Thus, LeNet-5 is identified as
CNN$_2$. We also consider batch normalization, which takes place after 
the dense or convolutional affine transformation, and before the
activation function, as is standard. We do not use biases
when we use batch normalization. We consider the widening factors,
\be
w \in \{ 2, 3, 4, 6, 8, 12 \}.
\ee

Finally, we consider a family of wide residual networks $\textrm{WRN}_w$, where $\textrm{WRN}_w$ is equivalent to table 1 of
\citet{WideRN} if $N=2$ and $k=w$. 
For consistency with \citet{WideRN} the first wide ResNet layer, a 3 to 16 channel expansion, is not scaled with $w$. 
As for CNNs, we study WRNs both with and without batch normalization.
We consider the widening factors,
\be
w \in \{ 2, 3, 4, 6, 8 \}.
\ee

The training sets of MNIST and Fashion-MNIST have been split into
training-validation sets of size 55000-5000 while for CIFAR-10,
the split is given by 45000-5000. We have used the official test set
of 10000 images for each dataset.
For MNIST and F-MNIST, we normalize the pixels of each image
to range from -0.5 to 0.5. For CIFAR-10, we normalize the pixels to have zero mean
and unit variance. We do not use data augmentation for any of the experiments presented in the main text.

\section{Training Time for Experiments}
\label{ap:ttime def}

For experiments with fixed learning rate $\epsilon_\textrm{fixed}$,
we set the number of training steps $T$ by setting both an
epoch bound $E_\textrm{min}$ and a step bound
$T_\textrm{min}$. So for a given batch size $B$
the number of training steps is set by,
\be\label{time bound}
T = \max \left(
T_\textrm{min}\,,~ E_\textrm{min} \cdot { N_\textrm{train} \over B}
\right).
\ee

After running the batch size search, we may choose a reasonable batch
size $B_\textrm{fixed}$ to hold fixed during learning search rate.
Experiments with fixed batch size and variable learning rate
are always paired to such a ``parent experiment."
When the batch size is fixed and the learning rate varies, we must scale the number of training steps proportional to the learning rate. We pick the reference learning rate $\epsilon_0$ to be the
learning rate at which the original batch size search was run,
and a reference number of training steps $T_0$, which is computed at the fixed batch size $B_\textrm{fixed}$ using
the epoch and step bound provided in equation \eq{time bound}. Then for learning rate $\epsilon$, the number of training steps $T$ is given by
\be
T = \max \left( T_0\,,~ T_0 \cdot {\epsilon_0 \over \epsilon} \right).
\ee
That is, for learning rates larger than $\epsilon_0$ we perform $T_0$ updates, while for learning rates smaller than $\epsilon_0$, we scale the number of updates inversely proportional to the learning rate.

\section{Experiment Details and Configurations}
\label{ap:configs}

In this section we detail the specific configurations of experiments run in this work.

\subsection{NTK without Batch Normalization}
\label{ss:nobx configs}

\begin{table}[h]
\caption{Epoch and step bounds for dataset-network family pairs
for various experimental settings.}
\label{t:nobx params}
\vskip 0.1in
\small
\begin{center}
\begin{tabular}{lllrrrrr}
\toprule
Dataset & Networks & $E_\textrm{min}$ & $T_\textrm{min}$
& $\epsilon_\textrm{BS\_search}$ & $B_\textrm{LR\_search}$ & $\epsilon_{0, \textrm{LR\_search}}$ & $T_{0, \textrm{LR\_search}}$ \\
\midrule
MNIST & 1LP & 120 & 80k & 10.0 & 8 & 10.0 & 825k \\
MNIST & 2LP & 120 & 80k & 10.0 & 16 & 10.0 & 412.5k\\
MNIST & 3LP & 120 & 80k & 10.0 & 16 & 10.0 & 412.5k\\
MNIST & CNN & 120 & 80k & 10.0 & 24 & 10.0 & 275k\\
F-MNIST & 1LP & 240 & 160k & 10.0 & 12 & 10.0 & 1100k \\
F-MNIST & 2LP & 240 & 160k & 10.0 & 24 & 10.0 & 550k \\
F-MNIST & 3LP & 240 & 160k & 10.0 & 48 & 10.0 & 275k \\
F-MNIST & CNN & 240 & 160k & 10.0 & 96 & 10.0 & 160k \\
CIFAR-10 & CNN & 540 & 320k & 5.0 & 256 & 10.0 & 160k \\
CIFAR-10 & WRN & 270 & 80k & 1.0 & 8 & 1.0 & 1500k \\
\bottomrule
\end{tabular}
\end{center}
\vskip -0.1in 
\end{table}

We run both batch size search and learning rate search to determine the optimal normalized noise scale
for networks trained with NTK parameterization and without batch normalization.
The relevant parameters used for the search experiments are listed in table \ref{t:nobx params}.
The scalar $\epsilon_\textrm{BS\_search}$ denotes the fixed learning rate used for batch size search,
while $B_\textrm{LR\_search}$ denotes the fixed batch size used during learning rate search.

The epoch bound $E_\textrm{min}$ and the training step bound $T_\textrm{min}$ are defined in section
\ref{ap:ttime def} of the supplementary material. Also as we explained in section \ref{ap:ttime def}, the training time is scaled with respect to a reference training time and a reference learning rate for the learning rate search experiments.
These are denoted $T_{0, \textrm{LR\_search}}$ and $\epsilon_{0, \textrm{LR\_search}}$ in the table
respectively.

\subsection{Standard without Batch Normalization}
\label{ss:nxbx configs}

\begin{table}[h]
\caption{Epoch and step bounds for dataset-network family pairs
for various experimental settings.}
\label{t:nxbx configs}
\vskip 0.1in
\small
\begin{center}
\begin{tabular}{llllrl}
\toprule
Dataset & Networks & $w$ & $E_\textrm{min}$ & $T_\textrm{min}$
& $\epsilon_\textrm{BS\_search}$ \\
\midrule
MNIST & MLP & All & 120 & 80k & 0.02 \\
F-MNIST & CNN & All & 480 & 320k & 0.03 \\
CIFAR-10 & WRN & 2, 3, 4 & 540 & 160k & 0.0025 \\
CIFAR-10 & WRN & 6, 8 & 1080 & 320k & 0.00125\\
\bottomrule
\end{tabular}
\end{center}
\vskip -0.1in 
\end{table}

For standard networks without batch normalization, we only carry out batch size search
experiments at a fixed learning rate $\epsilon_\textrm{BS\_search}$.
For CIFAR-10 experiments on wide residual networks, we chose to use two different
learning rates depending on the width of the networks (narrower networks can be trained faster with a bigger learning rate, while wider networks require a smaller learning rate for numerical stability).
The experiment configurations are listed in table \ref{t:nxbx configs}.

\subsection{NTK with Batch Normalization}
\label{ss:nobo configs}

\begin{table}[h]
\caption{Epoch and step bounds for dataset-network family pairs
for various experimental settings.}
\label{t:nobo configs}
\vskip 0.1in
\small
\begin{center}
\begin{tabular}{lllrlllr}
\toprule
Dataset & Networks & $E_\textrm{min}$ & $T_\textrm{min}$
& $\epsilon_\textrm{BS\_search}$ & $B_\textrm{LR\_search}$ & $\epsilon_{0, \textrm{LR\_search}}$ & $T_{0, \textrm{LR\_search}}$ \\
\midrule
MNIST & CNN & 120 & 80k & 10.0 & 128 & 10.0 & 80k \\
F-MNIST & CNN & 480 & 320k & 10.0 & 192 & 10.0 & 320k \\
CIFAR-10 & CNN & 540 & 320k & 10.0 & 192 & 10.0 & 320k \\
CIFAR-10 & WRN & 270 & 80k & 30.0 & 192 & 30.0 & 80k \\
\bottomrule
\end{tabular}
\end{center}
\vskip -0.1in 
\end{table}

The experiment configurations for networks parameterized using the NTK scheme
with batch normalization are listed in table \ref{t:nobo configs}.
Both batch size search and learning rate search have been carried out.
The parameters defined are equivalent to those used for NTK networks without batch normalization
in section \ref{ss:nobx configs}.

\section{Plots from Batch Search and Learning Rate Search}
\label{ap:plots}

In this section, we present plots of the average test set accuracy vs.
batch size/learning rate for batch size/learning rate search experiments
with fixed learning rate/batch size respectively. All the learning rate
search experiments are paired with batch size search experiments, and share
the same color-code and legend describing the network widening factors.

\begin{figure}[hp!]
  \centering
  \begin{tabular}{cc|cc}
  \includegraphics[height=2.8cm]{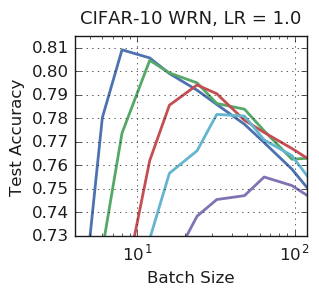} &
  \includegraphics[height=2.8cm]{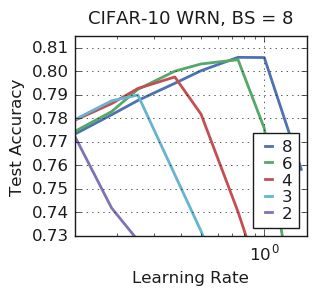} &
  \includegraphics[height=2.8cm]{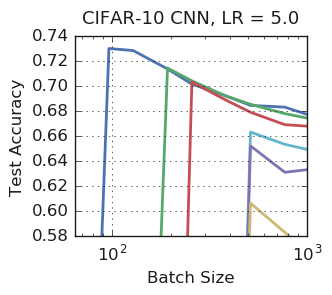} &
  \includegraphics[height=2.8cm]{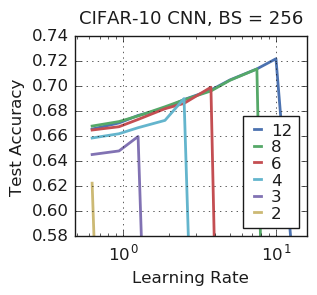} \\
  \includegraphics[height=2.8cm]{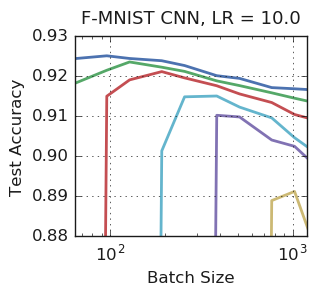} &
  \includegraphics[height=2.8cm]{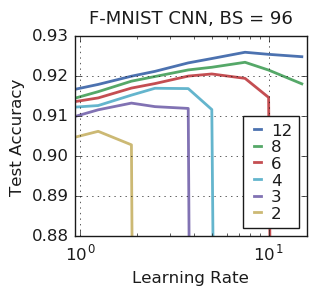} &
  \includegraphics[height=2.8cm]{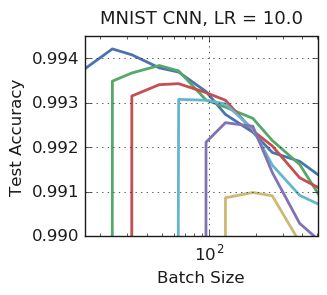} &
  \includegraphics[height=2.8cm]{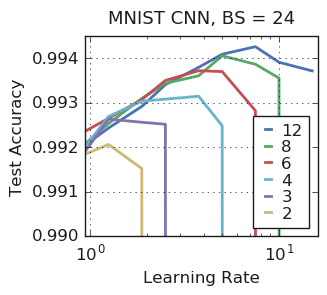} \\
  \includegraphics[height=2.8cm]{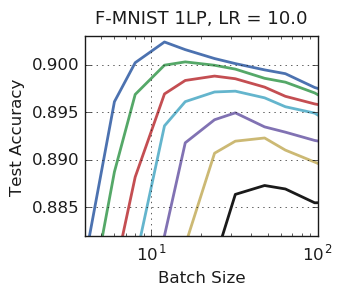} &
  \includegraphics[height=2.8cm]{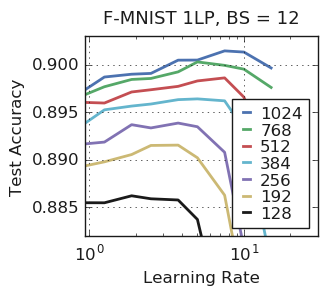} &
  \includegraphics[height=2.8cm]{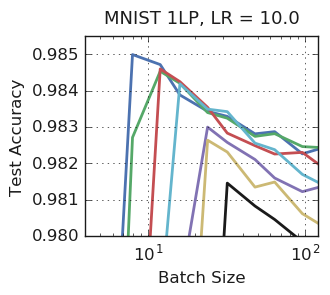} &
  \includegraphics[height=2.8cm]{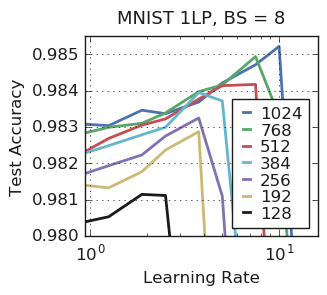} \\
  \includegraphics[height=2.8cm]{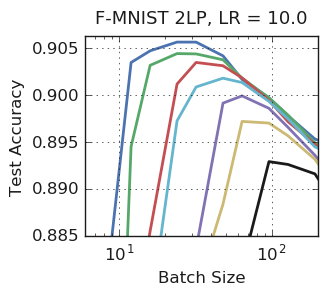} &
  \includegraphics[height=2.8cm]{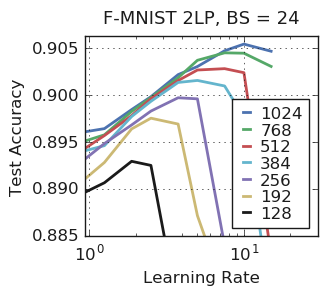} &
  \includegraphics[height=2.8cm]{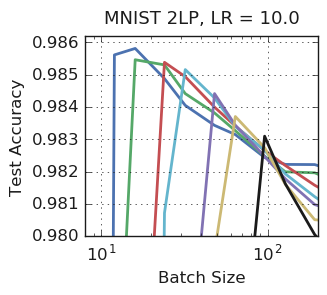} &
  \includegraphics[height=2.8cm]{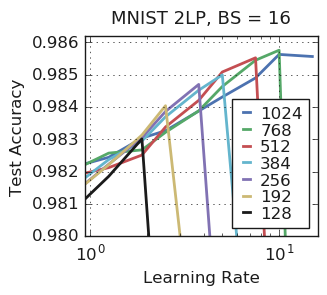} \\
  \includegraphics[height=2.8cm]{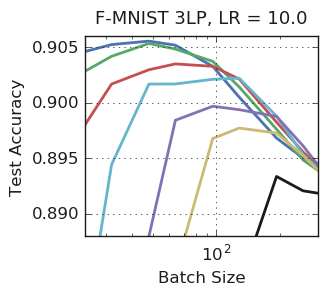} &
  \includegraphics[height=2.8cm]{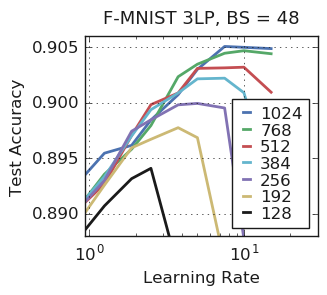} &
  \includegraphics[height=2.8cm]{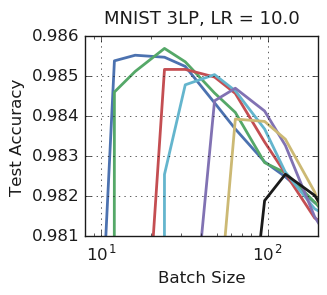} &
  \includegraphics[height=2.8cm]{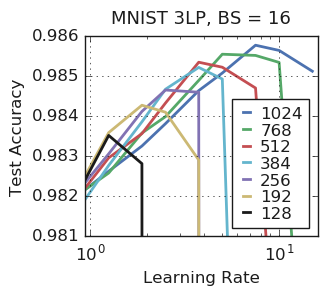}
  \end{tabular}
  \caption{The test accuracy plotted against batch size/learning rate
 for experiments at fixed learning rate/batch size respectively.
  These plots are from the experiments performed on NTK-parameterized networks
  without batch normalization. The legend indicates the widening factor
  of the plotted networks, and the x-axis is plotted in log scale.
  The values for the fixed parameters are indicated in the title.}
  \label{f:nobx search}
  \vskip -0.1in 
\end{figure}

Figure \ref{f:nobx search} plots the results of batch size/learning
rate search experiments run with NTK-parameterized
networks without batch normalization. Here, the x-axis is plotted
in log-scale. Since $\bar{g} \propto \epsilon/B$, if the
performance of the network is determined by the noise scale,
then the figures for batch size search and learning rate search
experiments on the same dataset-network pair should be symmetric to
one another.
This symmetry is nicely on display in figure \ref{f:nobx search}.

\begin{figure}[h!]
  \centering
  \begin{tabular}{ccc}
  \includegraphics[height=2.8cm]{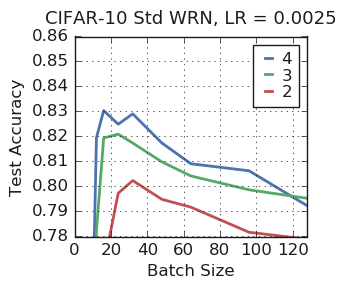}&
  \includegraphics[height=2.8cm]{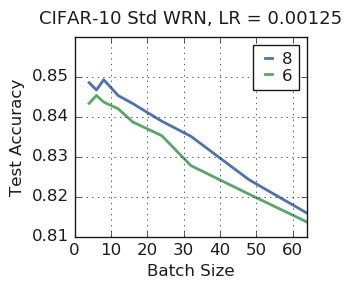}&
  \includegraphics[height=2.8cm]{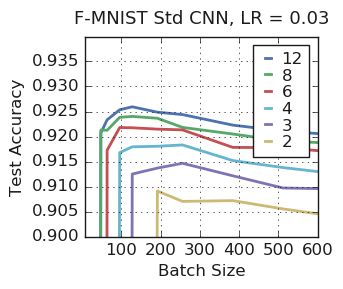} \\
  \includegraphics[height=2.8cm]{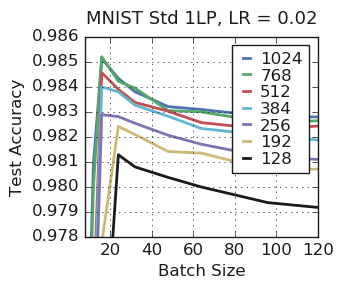} & 
  \includegraphics[height=2.8cm]{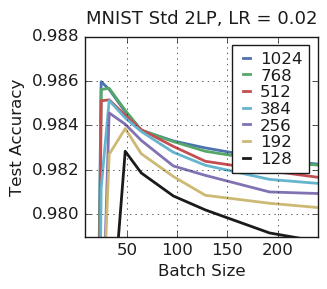} &
  \includegraphics[height=2.8cm]{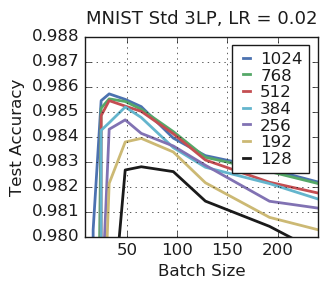} 
  \end{tabular}
  \vskip -0.1in
  \caption{The test accuracy plotted against batch size
  for experiments at fixed learning rate.
  The plots are from networks parameterized in the standard
  scheme without batch normalization. The legend indicates
  the widening factor of the plotted networks.}
  \label{f:nxbx search}
  \vskip -0.1in 
\end{figure}

\begin{figure}[h!]
  \centering
  \begin{tabular}{ccccc}
  \includegraphics[height=2.6cm]{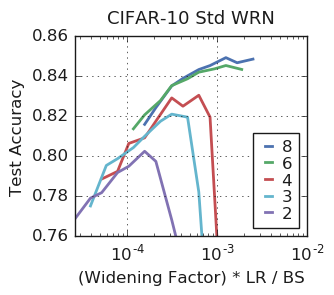}&
  \includegraphics[height=2.6cm]{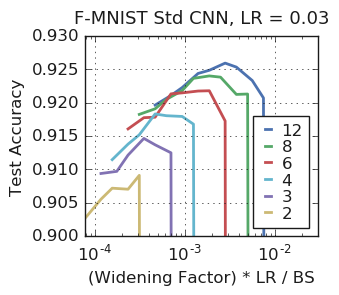} & 
  \includegraphics[height=2.6cm]{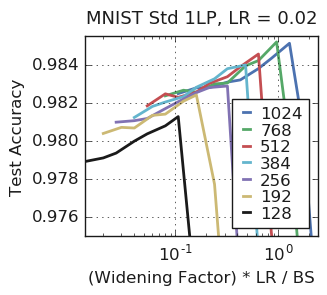} & 
  \includegraphics[height=2.6cm]{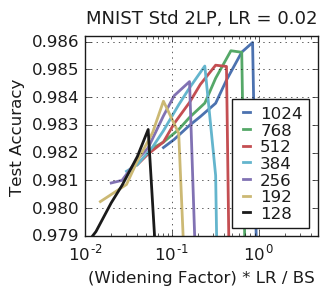} &
  \includegraphics[height=2.6cm]{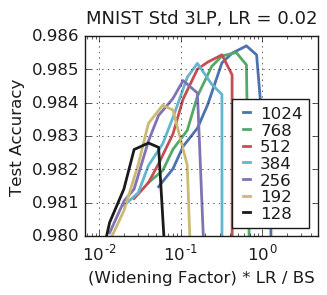} 
  \end{tabular}
  \vskip -0.1in
  \caption{The test accuracy plotted against $w \cdot \epsilon/B \propto \bar{g}$
  for networks parameterized in the standard scheme without batch-normalization.
  The x-axis is log-scaled. The legend indicates the widening factor.}
  \label{f:nxbx norm}
  \vskip -0.1in 
\end{figure}

Figure \ref{f:nxbx search}
plots batch size search experiments run with
standard parameterization and without batch normalization.
Standard-parameterized WRN$_6$ and WRN$_8$ were run with a reduced learning rate
due to numerical stability issues, and the results of
their batch size search experiments have been plotted separately. In contrast to NTK-parameterized networks,
the normalized noise is width-dependent for
networks parameterized using the standard scheme.
Also, we have batch searches conducted over varying learning rates in one
instance. Thus it is more informative to put everything
together and plot the performance of the network against
$(w\epsilon)/B \propto \bar{g}$. This has been done in figure \ref{f:nxbx norm}.
This plot reproduces the qualitative features of figure \ref{f:nobx search},
which is strong evidence for $\bar{g}$ being the correct
indicator of the performance of networks within a linear family.

\begin{figure}[h!]
  \centering
  \begin{tabular}{cc|cc}
  \includegraphics[height=2.8cm]{figures/bscan-c-wrn-nobo-log.png} &
  \includegraphics[height=2.8cm]{figures/lscan-c-wrn-nobo-log.png} &
  \includegraphics[height=2.8cm]{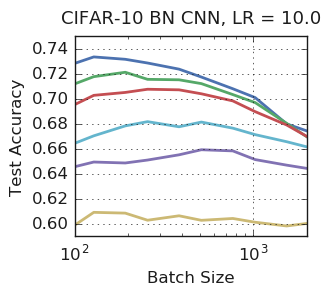} &
  \includegraphics[height=2.8cm]{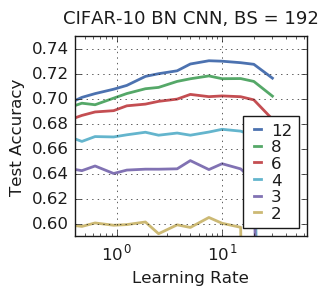} \\
  \includegraphics[height=2.8cm]{figures/bscan-f-cnn-nobo-log.png} &
  \includegraphics[height=2.8cm]{figures/lscan-f-cnn-nobo-log.png} &
  \includegraphics[height=2.8cm]{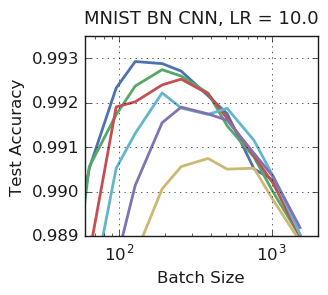} &
  \includegraphics[height=2.8cm]{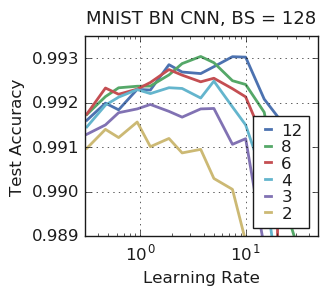} 
  \end{tabular}
  \vskip -0.1in
  \caption{The test accuracy plotted against (learning rate)
  for experiments at fixed learning rate/batch size respectively.
  The networks are NTK-parameterized
  and use batch normalization.
  The x-axis is log-scaled. The legend indicates the widening factor.}
  \label{f:nobo search}
  \vskip -0.1in
\end{figure}

Figure \ref{f:nobo search} plots the results of
batch size/learning rate search run with
NTK-parameterized networks with batch normalization.

When both batch size and learning rate search have been carried out,
the y-axes of the plots, along which the network performance is plotted,
are aligned so that the maximal performance obtained from the search
can be compared.

\section{Networks with Standard Parameterization and Batch Normalization}
\label{ap:stdbn}

In this section, we present results of batch search experiments with WRNs
with batch normalization that are parameterized using the standard scheme.
We train with a constant schedule with epoch bound $E_\textrm{min} = 270$
and step bound $T_\textrm{min}=80$k.

\begin{figure}[h!]
  \centering
  \begin{tabular}{ccc}
  \includegraphics[height=3.0cm]{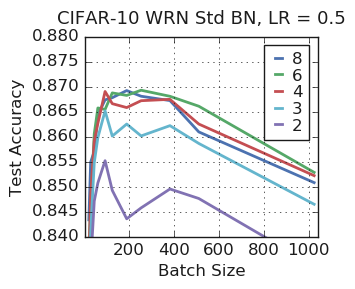}&
  \includegraphics[height=3.0cm]{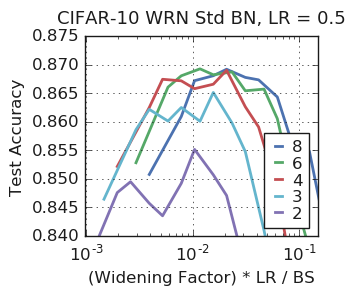} & 
  \includegraphics[height=3.0cm]{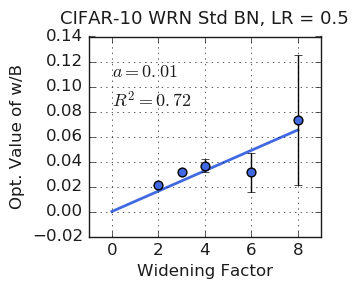} 
  \end{tabular}
  \vskip -0.1in
  \caption{Three plots summarizing the results of the batch search
  experiments with standard WRNs with batch normalization trained on
  CIFAR-10. The test accuracy of the networks
  are plotted against the batch sizes in the first figure, and 
  the value of $w \cdot \epsilon/B$ in the second figure.
  The x-axis is log-scaled in the second figure.
  The optimal value of $w/B$ is plotted against the network widths
  in the third figure.}
  \label{f:stdbnres}
 \vskip -0.1in 
\end{figure}

As was with the case with NTK parameterized WRNs, the scaling rule for
the optimum batch size coincides with that of the case when batch normalization
is absent, i.e., the optimal batch size ${B}_\textrm{opt}$ is constant with respect
to the widening factor.

\section{Batch Search Experiments with Regularization}
\label{ap:reg}

In this section, we present the results of training WRNs on CIFAR-10
with regularization. We use dropout with probability $0.3$
and label smoothing with uncertainty $0.9$.
Data augmentation is also applied by first taking a random crop of the
image padded by 4 pixels and then applying a random flip. Note that we have chosen these regularization schemes because we do not anticipate that the associated hyper-parameters will depend strongly on the network width or the noise scale.

\begin{figure}[h!]
  \centering
  \begin{tabular}{ccc}
  \includegraphics[height=3.0cm]{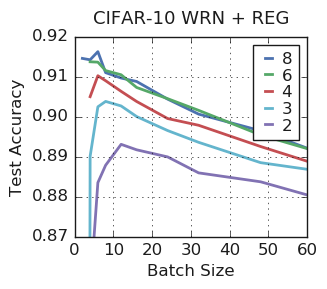}&
  \includegraphics[height=3.0cm]{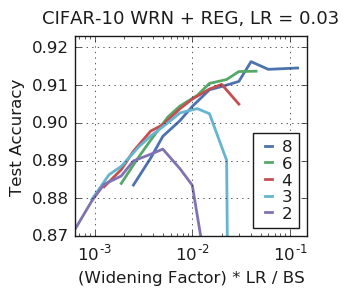} & 
  \includegraphics[height=3.0cm]{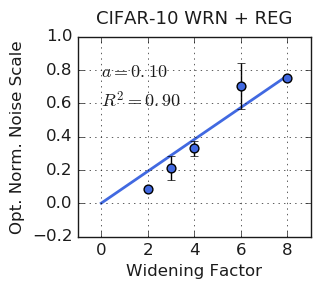} 
  \end{tabular}
  \vskip -0.1in
  \caption{Three plots summarizing the results of the batch search
  experiments with regularized WRNs trained on CIFAR-10.
  The test accuracy of the various networks
  are plotted against the scanned batch sizes in the first figure.
  In the second figure, the test accuracy is plotted against
  $w \cdot \epsilon/B \propto \bar{g}$ where the x-axis is log-scaled.
  The legend indicates the widening factor in the first two figures.
  The optimal normalized noise scale in units of $10^3$/[loss] is
  plotted against the network widths in the third figure.}
  \label{f:regres}
  \vskip -0.1in 
\end{figure}

We carry out batch search experiments for WRNs parameterized in the standard scheme
on CIFAR-10, with epoch bound $E_\textrm{min} = 270$ and step bound
$T_\textrm{min}=80$k. The results are given in figure \ref{f:regres}.
We have used the fixed learning rate $\epsilon=0.03$.

The scaling rule $\bar{g}_\textrm{opt} \propto w$ still holds
in the presence of these three regularizers. The use of regularization schemes significantly increases the final test accuracies, however these test accuracies still depend strongly on the SGD noise scale.

\section{The Performance of 3LP on a 2D Grid of Learning Rates and Batch Sizes}
\label{ap:grid}

\begin{figure}[h!]
  \centering
  \begin{tabular}{ccc}
  \includegraphics[height=3.0cm]{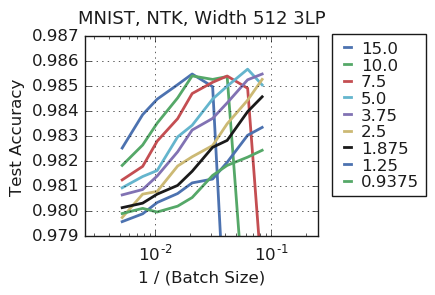}&
  ~~~~~ & 
  \includegraphics[height=3.0cm]{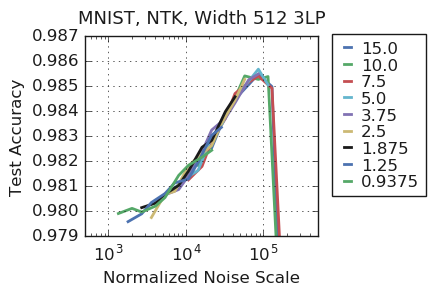} 
  \end{tabular}
  \vskip -0.1in
  \caption{The test set accuracy of a 3-layer perceptron (width 512) for a 2D grid of
  learning rates and batch sizes. We provide the learning rate in the legend and plot the test accuracy against the batch size (left) and the normalized noise scale (right).}
  \label{f:grid}
 \vskip -0.1in 
\end{figure}

Our main result (equation \ref{main result}) is based on the theory of small learning rate SGD, which claims that the final performance of models trained using SGD is controlled solely by the noise scale $g \propto \epsilon/B$.
To provide further evidence for this claim, here we consider 3LP$_{512}$ and measure its performance on MNIST
across a 2D grid of batch sizes $\{12, 16, 24, 32, 48, 64, 96, 128, 192 \}$ and a range of learning rates
as indicated in figure \ref{f:grid}.
We run 20 experiments for each learning rate/batch size pair and compute the mean test set accuracy.
We set the epoch bound $E_\textrm{min}$ to $120$ and the training set bound
$T_\textrm{min} = T_0 \cdot (\epsilon_0/\epsilon)$ with $T_0 = 412.5 \textrm{k}$ and
$\epsilon_0 = 10.0$. The results are shown in figure \ref{f:grid}, where
we plot the performance curves as a function of the batch size
and the noise scale. As expected, the final test accuracy is governed solely by the noise scale.

\newpage

\section{Numerical Stability and the Normalized Learning Rate}
\label{ap:stability}
In this section, we present experiments to verify the claim that
(in the absence of batch normalization), numerical instabilities
affect training when $\epsilon \gtrsim \epsilon_\textrm{unstable}$, where $\epsilon_\textrm{unstable}$ is constant with
respect to the width for NTK parameterization while
$\epsilon_\textrm{unstable} \propto 1/w$ as $w \rightarrow \infty$ for networks parameterized using the standard scheme.
This is equivalent to saying that the scale $\bar{\epsilon}_\textrm{unstable}$
at which the normalized learning rate becomes unstable
is constant with respect to the width of the network.

To do so, we take families of NTK and standard parameterized networks,
and compute the failure rate after 20 epochs of training
with a fixed batch size (64) at a range of learning rates.
For each network, we run 20 experiments, and compute the failure rate,
which is defined to be the portion of experiments terminated by a numerical
error. We run the experiments for CIFAR-10 on WRN$_w$ ($w=2, 3, 4, 6, 8$),
F-MNIST on CNN$_w$ ($w=6, 8, 12, 16, 24$)
and MNIST on 2LP$_{512 \cdot w}$ ($w=1, 2, 4, 8, 16, 32$).

For CNNs and 2-layer perceptrons, we consider much wider networks
than are studied in the main text. This is because we are ultimately
interested in observing numerical instabilities which occur when $w$ is large. For the purpose of studying this break-down of numerical stability, we can afford to use much wider networks. The width dependence of
$\epsilon_\textrm{unstable}$ becomes more evident by focusing on these wide networks, as the behaviour of narrow networks is less predictable.

\begin{figure}[b!]
  \vskip -0.1in
  \centering
  \begin{tabular}{cc|c}
  \includegraphics[height=3.7cm]{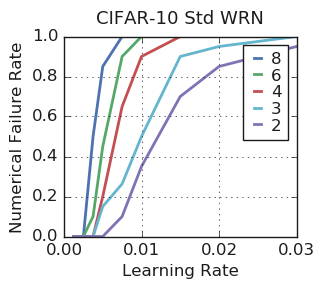} &
  \includegraphics[height=3.7cm]{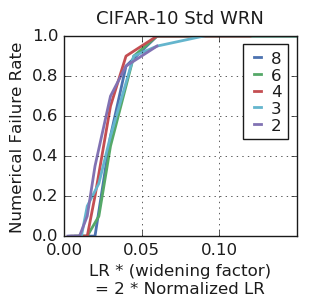} &
  \includegraphics[height=3.7cm]{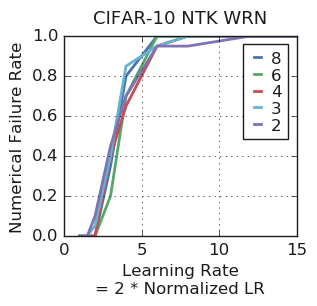} \\
  \includegraphics[height=3.7cm]{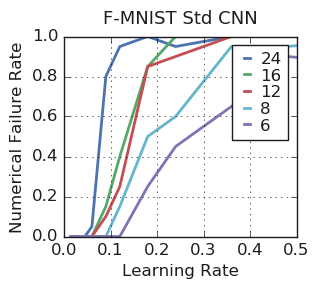} &
  \includegraphics[height=3.7cm]{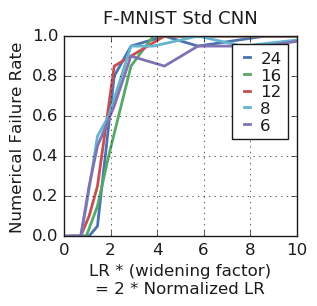} &
  \includegraphics[height=3.7cm]{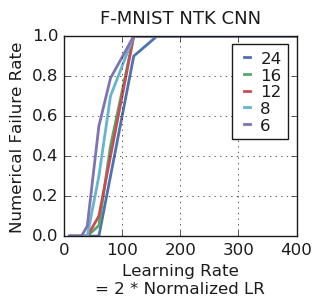} \\
  \includegraphics[height=3.7cm]{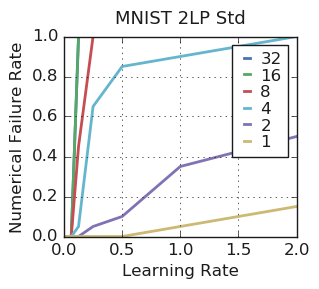} &
  \includegraphics[height=3.7cm]{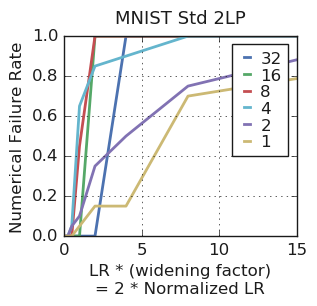} &
  \includegraphics[height=3.7cm]{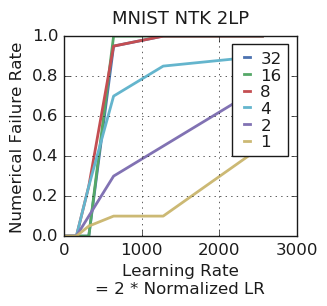}
  \end{tabular}
  \vskip -0.1in
  \caption{Failure rate plotted against the learning rate or the normalized learning rate
  for various dataset-network pairs.}
  \label{f:stability}
  \vskip -0.1in 
\end{figure}

Figure \ref{f:stability} depicts 3 figures for each dataset-network combination.
The first figure shows the failure rate plotted against the learning rate for networks using
the standard parameterization. The second is the failure rate plotted against the product
of the learning rate and the widening factor---i.e., twice the normalized learning rate---for
the same networks (trained using the standard parameterization). The third figure shows the failure rate
plotted against the learning rate for networks parameterized using the NTK scheme. Here, the normalized
learning rate is simply half the learning rate.

It is clear from these plots that $\bar{\epsilon}_\textrm{unstable}$ is independent of
the widening factor for WRNs and CNNs, while the definition of
$\bar{\epsilon}_\textrm{unstable}$ for the 2LP seems more subtle.
Nevertheless for all three network families, we see similar stability curves as width increases for both parameterization schemes, when measured as a function of the normalized learning rate.

\newpage

\section{Performance Comparison between NTK Networks and Standard Networks}
\label{ap:ntk vs std}

\begin{figure}[h!]
  \vskip 0.15in
  \centering
  \begin{tabular}{cccc}
  \includegraphics[height=3.7cm]{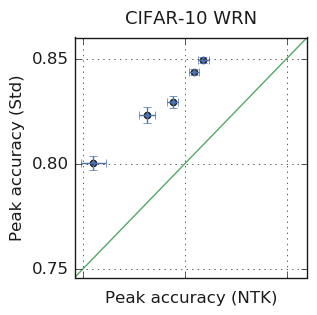} &
  \includegraphics[height=3.7cm]{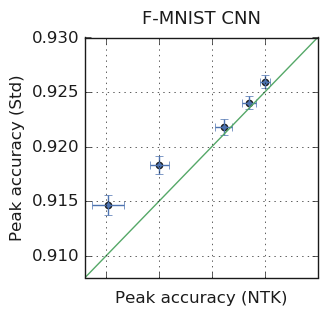} &
  \includegraphics[height=3.7cm]{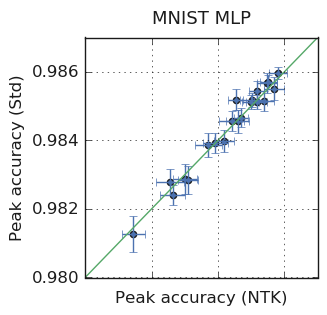} 
  \end{tabular}
  \vskip -0.1in
  \caption{Performance of networks with standard parameterization plotted against
  performance of networks with NTK parameterization. The green line is the line $y=x$.}
  \label{f:ntk vs std}
  \vskip -0.1in 
\end{figure}

Here we provide a brief comparison of the performance of both parameterization schemes
on the test set.
In figure \ref{f:ntk vs std}, the peak test accuracy of a network parameterized
with the standard scheme is plotted against the peak test accuracy obtained when
the same network is parameterized with the NTK scheme. The dataset-network pairs
are indicated in the title---CIFAR-10 on WRN$_w$, F-MNIST on CNN$_w$ and MNIST on MLPs.
We see that standard parameterization consistently out-performs
NTK parameterization on WRNs and CNNs, although the performance is comparable
for MNIST on MLPs.

\begin{figure}[h!]
  \vskip 0.15in
  \centering
  \begin{tabular}{ccc} 
  \includegraphics[height=3.7cm]{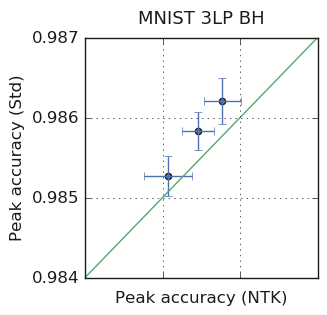} &
  ~~~~~ &
  \includegraphics[height=3.7cm]{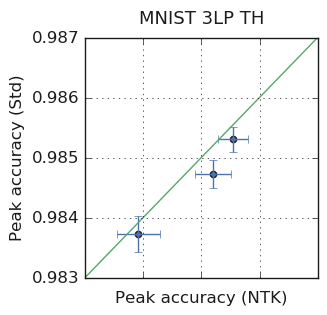} 
  \end{tabular}
  \vskip -0.1in
  \caption{Performance of networks with standard parameterization plotted against
  performance of networks with NTK parameterization for 3LP networks BH$_w$ and TH$_w$. The green line is the line $y=x$. }
  \label{f:ntk vs std MLP}
  \vskip -0.1in 
\end{figure}

The reason that performance agrees well for the particular MLPs investigated in the
main text of the paper, is because all the hidden layers have equal width. In this limit, NTK parameterization and standard parameterization are essentially identical.
However by varying the network architecture, we can observe a discrepancy between
the performance of MLPs as well. As an example, we consider the following
bottom-heavy (BH) and top-heavy (TH) 3LPs with hidden layer widths,
\be
\textrm{BH}_w~: \quad [4w,\,2w,\,w]\,, \qquad
\textrm{TH}_w~: \quad [w,\,4w,\,4w].
\ee
We consider the widening factors $w \in \{128,256,512\}$.
In figure \ref{f:ntk vs std MLP}, we display the discrepancy between the performance
between both bottom-heavy and top-heavy networks parameterized in the NTK and standard schemes.

In the bottom heavy case, MLPs parameterized in the standard scheme appear to outperform MLPs parameterized in the NTK scheme. However in the top heavy case, MLPs parameterized in the NTK scheme appear to outperform MLPs parameterized in the standard scheme. These results suggest that neither scheme is superior to the other, but that the final performance will depend on the combination of parameterization scheme, initialization conditions and network architecture. We note that we initialize all our networks at critical initialization ($\sigma_0^2 = 2$), and that this overall weight scale was not tuned.

\end{document}